# Automatic Transcription of 17th Century English into Contemporary English with NooJ: Method and Evaluation

Odile Piton, SAMM, Paris-1 University, France,

Slim Mesfar, RIADI, Manouba University, Tunisia,

Hélène Pignot, SAMM, Paris-1 University, France

## Introduction & Motivations

The treatment of ancient texts by numerical methods (statistics & NLP) develops widely, thanks to the creation of new tools and applications. For instance, the LAMOP laboratory at Panthéon-Sorbonne University, among many other research interests, is involved in computer-assisted studies of Latin, French and English medieval texts. There are also two research units worldwide that study variation and change in historical corpora; one is based in Lancaster and the other in Helsinki. At Helsinki University the VARIENG research unit includes 50 members. Focusing on empirical research, "they compile and use electronic corpora of English, and develop tools and methods for corpus-based and ethnographic studies". They work on variation in historical corpora, but to our knowledge they do not provide any transcription tools (Andersen 2011).

At Lancaster University, a research team first designed a tool called VARD to locate spelling variants in early modern English (from 1450 to 1700) and provide their equivalents in modern English. But this tool, which was limited to a predefined manually compiled list, could not account for all possible variations. Version N° 2 (Baron 2008) uses replacement rules for certain letters (c for ck, delete the final silent e, replace 'd by ed, for example) and windows open displaying each of the proposed equivalents, and the probability rate that a particular transcribed form is the right word or expression in contemporary English.

Our earlier research focused on corpora of 17th Century English travel literature and aimed at describing and processing with the linguistic platform NooJ the spelling variants and morpho-syntactic structures specific to 17th century English (Pignot 2010, 2011, Piton 2009). In this article, we are using the results of previous research and proposing a transcription method for words or sequences identified as archaic.

To that end, new tools—using the TRANS operator and new functionalities developed by M. Silberztein (Silberztein 2006), especially in 2006 (Mesfar 2006)—have been designed to provide the necessary environment producing the automatic transcription of 17th century English into contemporary English. This essay will therefore present our methodology, spell out the rule order that was implemented, and finally compare the original experimentation corpus with the transcriptions suggested in the text annotation structure, to test the performance of these new tools and evaluate their success rate.

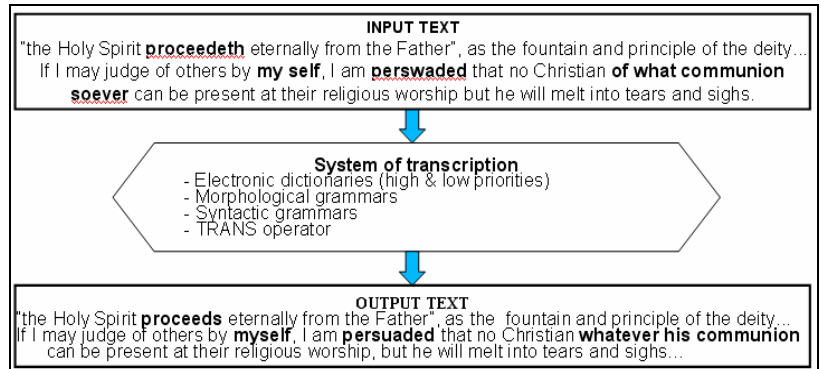

**Figure 1.** General Model

## Methodology

Our methodology relies on recognition tools and transcription operations. Recognition tools comprise electronic high & low priorities dictionaries, morphological grammars and syntactic local grammars. Transcription operations are divided into five main classes: word for word transcription; concatenation of disjointed forms; split-up of contractions; "syntagm for paradigm" transcription and syntactic transformation.

In some cases, these 5 transcription classes may need to be combined. Some difficulties arose, which will be detailed and explained further down.

The suggested transcriptions are propositions that require user validation. Most of the time our graphs or dictionaries suggest only one transcription, but in some cases no correct transcription or several transcriptions with only one correct answer. It is up to the user to choose the good one. A numerical assessment of their accuracy will be produced at the end of this paper.

## 1st Transcription Class: "Word for Word Transcription"

The first transcription class deals with archaic forms and spelling variants. For the transcriptions of archaic forms, electronic dictionaries with high or ordinary priorities are used. For spelling variants, potential transformations need to be explored. We have described substitutions, insertions or deletions of letters, including doubling consonants and adding silent "e"s. The transformation is carried out by the TRANS operator while the CE form is described by the formula EN=CElemma in a dictionary entry or sometimes in an entry built by a morphological graph. The recognised forms will give the syntactical TRANS operator the necessary clues to perform the transcription (Silberztein 2007). The annotated text can be exported as an XML document.

> For an invariable word: ArchForm,CATEGORY+ EN=CElemma
> For a verb it is:ArchForm,V+FLX=VerbPARADIGM[1]+EN=CElemma.
> For a noun it is:ArchForm,N+FLX=NounPARADIGM+EN=CElemma.
> E.G.: unlesse,CONJ+EN="unless"; stile,V+FLX=SMILE+EN="style"; burthen,N+FLX=Nsp+EN="burden"; bisquet,N+FLX=Nsp+EN="biscuit"; chyrurgion,N+FLX=Nsp+EN="surgeon"; betimes,ADV+EN="early".

We assume that forms to be transcribed are stored in a NooJ variable designated as $W. The TRANS operator uses the formula $W$EN to obtain the corresponding CE lemma. For invariable words this CE lemma is the CE form. Concerning inflected words, we need to be more specific. The TRANS operator has a path for each inflexion. For a noun, there is one path for the singular (+s) and another for the plural (+p). For a noun in the plural the path is ($W<N+p+EN>)/<TRANS+RES= $W$EN_N+p>) where <N+p+EN> means that the path deals with a noun (N) in the plural (p) that has the feature EN (EN=CElemma). We insist on the surrounding naming expression ($W). The TRANS operator will store the result of $W$EN_N+p in the RES property. For example, if we consider the word "bisquets", $W$EN will give the lemma biscuit, then from this lemma the formula _N+p gives the plural form of the lemma biscuit, thus, it produces the output "biscuits".

---

[1]FLX=XXX shows the model to build inflected forms for a noun or a verb, from a model named XXX, there are as many inflexion models as needed.

The transformation of verbs must take into account the tense, the number as well as the person. We can treat spelling variants and flexion variation in verbs, with different final morphemes as in "saith" (says), or "dippeth" (dips). The second person in the present is inflected in -est and the third person could be spelled -eth or -th instead of -s or -es. Morphological grammars recognize potential transformations compared with lexical entries stored in electronic dictionaries. When syntactic grammars are applied, they produce the transcription.

> For example the form "liveth" is recognised and described as liveth,live,V+Tense=PR+3+Nb=s+EN=live. This refers to the equivalent form "lives" of the lemma "live" that the TRANS operator creates automatically, following a transcription path designed for the third person singular present: ($W<V+PR+3+s+EN>)/<TRANS+RES=$W$EN_V+ PR+3+s>. As for nouns, "$W$EN" refers to the lemma; however, "_V+PR+3+s" indicates that the third person singular present is required.

The second transformation type processes the preterit and the present perfect, which could be shortened to t or d as seen further down, or 't or 'd as stated in the next section. For instance, in our text we may single out the forms linkt (linked), dismist (dismissed), and cropt (*i.e.* cropped, one of the doubled consonants being omitted).

> These forms are processed by the TRANS operator thanks to the following path that will produce the CE equivalents: ($W<V+PP+EN>)/ <TRANS+RES=$W$EN_V+PP> where V+PP refers to the past participle.

## 2nd Transcription Class: "Concatenate"

The forms to be processed include disjointed words, hyphenated compound words and past participles that were spelled 'd (or 't).

There were different occurrences in the experimentation text: e.g. my self, no body, back wardness, to day, and any thing. The method is the same as in the previous class since dictionary entries can register compound words including a space as well as a hyphen. In so far as the word is correctly identified in the dictionary, the entry indicates the archaic form and the new form, and the TRANS operator will automatically generate the form in CE. This is very similar to the first transcription class.

> "my self" is thus described as: my self,PRO+EN="myself", "where-ever" is described as where-ever,PRO+EN="wherever" or where-ever,CONJ +EN="wherever". The transcription of invariable words is produced for pronouns by the path: ($W<PRO+EN>)/<TRANS+RES=$W$EN>.

Spelling variants are taken into account in specific entries, e.g. my selfe,PRO +EN="myself".

But here come the differences. The first transcription class cannot deal with unrecognized disjointed tokens since morphological graphs can process one token only. For separate unrecognized forms which morphological graphs are unable to handle, we must introduce other tools able to deal with them and build new "annotations" dynamically. They are "syntactical graphs designed to work on morphology". They are saved, as other syntactical graphs, in the directory named Syntactical Analysis.

Let's now direct our attention to elided past participles spelled 'd or 't (some nouns and adjectives spelled this way may also be recognised, as we shall see further down), hence forms like "allow'd", "profes'd" (with the elision of two letters), "dry'd" (elision and substitution of final letters), "joyn'd" (substitution of one letter), "imbrac'd" (in which two combined morphological transformations are required to produce its CE equivalent "embraced"). The dictionary entry plays a key role in the recognition process, but the identification of the form is difficult when the word is broken down into two tokens, and the first is not necessarily analysable as a verb. So the results of the annotation step are discarded.

In the four examples we took, "allow" and "dry" are recognized as verbs in the infinitive (renamed $W) whose past participle (_V+PP) will be generated by the TRANS operator without further ado with the path ($W<V+INF-EN>)'d/<TRANS+RES=$W_V+PP>. Joyn and imbrac(e) are archaic spellings of the verbs embrace and join. They are registered as CE lemmas for imbrac and join. $W$EN refers to the CE lemma embrace vs join while _V+PP refers to the past participle of the lemma. In a word, imbrac'd is correctly processed by the TRANS operator which produces the past participle of embrace as the correct output of ($W<V+INF+EN>)'d/<TRANS+RES= $W$EN_V+PP>.

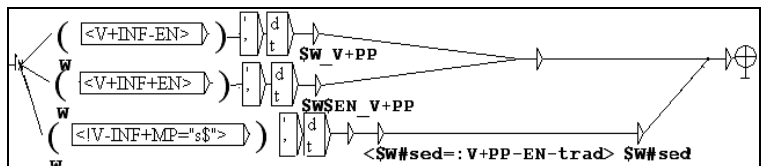

**Figure 2.** Some paths in the transcription graph of forms spelled 'd or 't

What about nouns and adjectives? It is a well-known property of these word classes that they may be turned into compounds by adding the -ed inflection to their form. In our corpus we found only one occurrence of this: "the long-hair'd Greeks". We thus devised our tool in a way that it can also operate the transcription of adjectives and nouns in 'd.

Some transcriptions of words wrongly recognised as nouns are proposed. Because of polycategory, some words may be recognized as verbs, but also as nouns and adjectives, which may result in erroneous transcriptions. Propositions need to be validated and incorrect answers filtered out.

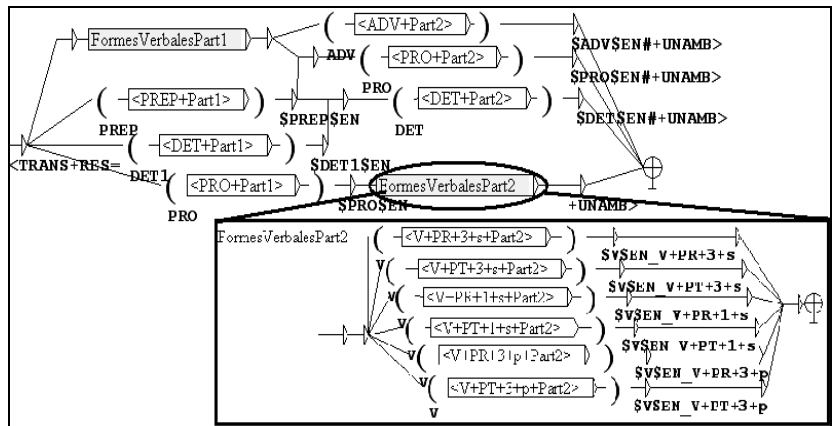

**Figure 3.** Syntactical grammar: contractions and details of Part2.

## 3rd Transcription Class: "Split up Contractions"

Let's give a few examples of contractions (some are still used in poetry): "'tis" for "it is", "t'other" for "the other", "ith'" for "in the". The transcription of contractions is processed thanks to two specific tools: a high priority dictionary with the lexical annotations describing the two parts of contractions, named Part1 and Part2, and a syntactic grammar.

> For instance, "'tis" is described as: 'tis,<it,it,PRO+3+n+s+Part1+EN=it> <is,be,V+PR+3+s+Part2+EN=be>+UNAMB. In other words, " 'tis" is made up of the third person pronoun "it" followed by the third person verb be in the present. This description is rounded off with the property UNAMB which rules out all other interpretations. Likewise, "'twas" is described as the preterit form PT: 'twas,<it,it,PRO+3+n+s+Part1+EN=it> <was,be,V+PT +3+s+Part2+EN= be> +UNAMB.

The second tool to be used is a syntactic grammar dedicated to generating the transcription of the two parts of the contraction. This grammar, which analyses the components of the contraction and produces the adequate forms, is presented in Figure 3.

## 4[th] Transcription Class:"Syntagm for Paradigm"

Let us now examine how NooJ transcribes forms whose equivalent in CE may be a syntagm. In this case a single paradigm may be replaced by one expression only and this is indicated by the REPLACE attribute. The transcription is processed using the formula $W$REPLACE.

> The adverb or conjunction whencesoever may be transcribed as "from whatever place", which results in two entries:
> whencesoever,ADV+REPLACE="from whatever place"
> whencesoever,CONJ+REPLACE="from whatever place".

We then process words that will remain intact in the text but about which we want to provide information thanks to a text in parentheses. The text is indicated by the NOTE trait.

> For the entries "pix" and "saique" are described as:
> pix,N+FLX=Nsp_es,+NOTE="a box where the Holy Communion is kept"
> saique,N+FLX=Nsp+NOTE="a big bark".

Thirdly some words must be transcribed as expressions whose components include inflected nouns or verbs. Verbs or nouns need to be transcribed with the appropriate inflection, and they can be preceded and/or followed by an invariable word or phrase. It might be a verb preceded by an adverb or a noun preceded by an adjective, or a verb followed by a complement. All these cases may be treated and inflected forms preceded or followed by an invariable term can be transcribed. To do so, we have added special properties to the dictionary, *i.e.* PREINSERT and POSTINSERT. The preceding form1 is indicated by PREINSERT=form1, and the following form2 is indicated thanks to the code POSTINSERT=form2.

> Acquiesce, acquiesces, acquiesced, and acquiescing, hence all of the inflected forms of the archaic verb acquiesce ("remain at rest"), may be transcribed into "remain at rest", "remains at rest", "remained at rest", "remaining at rest" thanks to the following dictionary entry:
> acquiesce,V+FLX=LIVE+EN=remain+POSTINSERT="at rest"

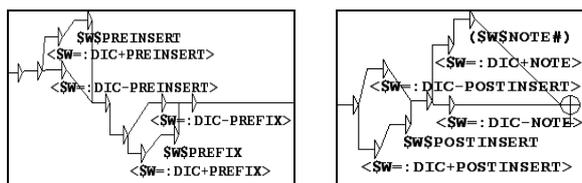

**Figure 4.** Syntactic grammars for PREINSERT, POSTINSERT and PREFIX

> The word "accoustrement" may be replaced by "accessory item of clothing". The noun item is transcribed in the singular or in the plural, the adjective accessory is anteposed, and the complement "of clothing" will be postposed. The entry is accoustrement,N+FLX=Nsp+PREINSERT= "accessory"+EN="item"+POSTINSERT="of clothing".

Figure 4 shows the sections of the graphs using PREINSERT and POSTINSERT. We have also created the PREFIX attribute that can add a prefix to a word to create a new one. There is a slight difference between a PREFIX and the PREINSERT attributes: A "pre-inserted" form is followed by a space whereas a prefix is not, which allows for its concatenation with the next word on the string.

> Belov'd is recognized as a prefixed form of loved and a graph suggests this entry: belov,V+INF+PREFIX=be+EN=love.

## 5th Transcription Class: Syntactic Transformation

Some archaic sequences need to be rewritten. We have been able to treat certain sequences such as the expletive or periphrastic "do" (he does believe=he believes, he did believe=he believed, they do believe=they believe), the forms in "soever" ("no Christian of what confession soever"), and the use of which as a relative pronoun when the possessor is human. Each of these forms requires a local grammar to rewrite them. Some elements in the forms in "soever" need to be permuted. The transcription tool developed with NooJ may locate the components that need permutation, those that do not vary and those needing transformation. After recognising the pattern, our local grammar generates the matching sequence by transformational analysis. This is done by the graph in Figure 5. Thus "no Christian of what communion soever" becomes "no Christian whatever his communion"; "of what nation or quality soever" becomes "whatever his nation or quality"; "how strict soever" becomes "however strict".

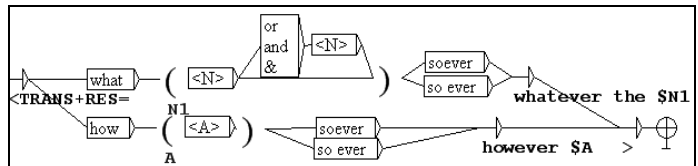
**Figure 5.** Graph processing some forms in "soever"

Some transformations are so complex that the transcription process must be repeated; the first run produces a word for word transcription of simple words thanks to grammars, but the second run may identify grammatical sequences that were not recognised before, because there are two close syntactical transformations in the sentence.

If we have a past participle spelled 'd as in "dismay'd", within a sequence like "how dismay'd soever this woman is", our local grammar cannot transcribe it in only one run; the first transcription transforms dismay'd into dismayed, and the second one will perform the final transcription "however dismayed this woman is".

## Evaluation and Discussion

Traditionally, the scoring report compares the generated transcribed text file with a carefully annotated file. The system was evaluated in terms of the complementary precision (P) and recall (R) metrics. Briefly, precision evaluates the noise of a system while recall evaluates its coverage. These metrics are often combined using a weighted harmonic called the F-measure (F).

**P= # of correct transcriptions/ # of automatically transcribed entities**
**R= # of correct transcriptions / # of entities to be transcribed**
**F = 2 P R / P+R**

The evaluation is carried out on our test corpus of English travel literature. This corpus contains around 9000 word forms with about 2600 different tokens. Our first remark is that 15% (387 tokens) of this list of tokens tagged as 17$^{th}$ century English. The experimentation gives the following scores:

Precision = 96.85%
Recall = 94.36%
F-Measure = 95.58%

Most words in our small test corpus are correctly transcribed. However, we notice that 11% of the correctly transcribed words or sequences were ambiguous especially because of their initial analysis with more than one part of speech. This problem will be solved by means of disambiguation grammars and will be described in a further work.

In addition, we observed two remaining main problems that are not technically solved yet: proper nouns whose spelling was modified were transcribed, but without a capital letter. For example, "Romane" was transcribed as "roman".

Some competing graphs suggested up to three transcriptions for some short paronymic words: for instance for the word "sinne" three transcriptions were provided, "sine", "sin" and "sines". For the word "poore" the transcription "pore" was proposed as well as the correct one. When there is no paronymic word close to the word requiring transcription, the correct transcription is given: for example "nunnes" was correctly transcribed as nuns (there is no form or entry like "nune" in CE).

Moreover we must mention the problem of words whose flexion was different from CE, adjectives ending in -ese could be pluralised (hence "Chineses") and some words such as information were countable. To treat these cases the only solution to get a correct transcription is to add these entries to the higher priority dictionary with a relevant flexion.

Lastly there was some silence for words whose spelling is too modified: words like "fatned" or toung (in another corpus) are tagged as unknown. Dictionary entries are required in order to handle this type of case.

## Morphological versus Syntactical Graphs and Hierarchy

Dictionaries and graphs have a priority level, and each word is searched from the highest level (*i.e.* with the highest priority) to the lowest. Each level is thoroughly explored, if the word is not found at one level, the research is carried on to the next one. When it is successful, the search stops when the whole level has been explored.

Consequently, the choice of priorities for morphological graphs and dictionaries is important. Graphs that recognise certain parts in words (the beginning or the end of a word) are given a higher priority over those that recognise medial variations. A lower level is given to graphs that deal with prefixed words. An irrelevant choice of priorities could produce wrong results only. This order is data-driven and empirical, and may still be improved after further testing.

Let us give some precisions about levels. Some graphs are in the middle (default) level, such as the graph processing Roman numerals. However, most of them are assigned a lower priority (we are using three levels of low priority) and appealed to only when the words to be transcribed do not feature in any of the dictionaries. We are using three dictionaries, the NooJ Sdic dictionary (for contemporary English), S17dic our dictionary of 17$^{th}$ century English with validated forms as well as Prioritydic which is a dictionary with a higher priority[2] to recognize specific words that NooJ may wrongly recognise thanks to Sdic. The distribution between dictionaries must be made carefully.

Syntactical graphs are applied after dictionaries and morphological graphs. We run the disambiguation graphs first and this filters out some of the wrong lemmas.

## Some Remarks

In this article we have not tackled the question of traits in the dictionaries. The file Properties.def lists "categories and properties associated with features". For nouns in Sdic we have several features coded Abst (abstract), Anl (animal), AnlColl (animal + collective), Conc (concrete), Hum (human) etc. We add properties such as XVII or Priority to our own dictionaries.

Sometimes we may be misled by our knowledge of and intuitions about the lexicon. Let us give an example. In our text we come across the word "knowes" that off the top of our heads seems an archaic spelling for "knows". Curiously enough, it has been falsely recognised as a noun by Sdic. Without it, one of our graphs would have found the accurate annotation. The reason is that Sdic contains some rare or dialectal words. "Knowe" is registered as a noun. It is a Scottish form for "knoll". This can be intriguing. Disambiguation tools are inefficient in such a case. They are designed to choose between several annotations, discard the irrelevant ones and keep the exact one, but for "knowes", there is no choice because there is only one irrelevant annotation produced from a direct dictionary lookup.

The transcription is foreseeable as long as the entry is correctly described, which brings us back to the identification step, the trickier of the two as it needs to work out the necessary morphological transformations to be performed.

Our second remark is triggered off by the fact that morphological graphs are working on a single word (more precisely a single token) and not a sequence of words. The token is treated out of context. Our graphs automatically perform multiple transformations. For each token treated by our graphs, one form per path in every graph of the search level in progress is tested, then the annotation is produced, along with a suggested transcription. If no word is found, then the search moves on to the next inferior level. Knowing the context, the reader spontaneously eliminates much of them as nonsense. He should not forget that this excess result is a drawback of automatic treatment.

---

2 As M. Silberztein says in the NooJ manual: "this small higher priority dictionary will hide useless Sdic entries and act as a filter, to filter out unwanted entries".

## A few examples

Let us detail how we deal with words that had different meanings or different grammatical classes.

In CE the word "penitentiary" is a noun and an adjective. As a noun it means a "jail". In the 17th century the noun "penitentiary" meant a "penitent", or a "spiritual father". It could be an adjective too. To describe these transcriptions we have two entries: penitentiary,N+FLX=Nsp_y+ PRE=spiritual+EN=father and penitentiary,N+FLX=Nsp_y+ EN=penitent. In order to hide the CE entry as a noun, we must register the two new entries in Prioritydic, and add the adjective entry: penitentiary,A. They will hide the corresponding entries of Sdic.

The word vassal could be either a noun (as usual) or a verb (arch. = to submit). In this case we have nothing to hide, only a category to add. So we will describe vassal as a verb in an entry of S17dic.

A word like "putteth" is an archaic form of the third person present "puts". We created a graph that transforms a verb ending in -eth or -th into -es or -s. So this graph can generate the form putts as "putt" features in Sdic. This CE dictionary recognises the verb "putt" being a golf term attested as early as 1743. This is unsuitable diachronically. Therefore we need to "hide" this entry. Our aim is to prevent NooJ from recognizing "putt" as a golf term and the solution we found is to insert the new entry indicating the transformation into Prioritydic, this way NooJ will locate it and not look into a lower level dictionary (here Sdic): putt,V+FLX=HELP+EN=put.

Whenever a form or a phrase specific to 17th English is found, one transcription (and sometimes more) is suggested for each of them. It is up to the user to choose the correct one. The selected form will be included in the transcribed text.

## Conclusion

Some transformations are tricky, for instance identifying when one should replace the relative pronoun "which" by who (in 17$^{th}$ century English the pronoun "which" could be used when the antecedent is human). We have not yet processed inversions of the subject and the postposition of adjectives. Their recognition is very difficult because of the phenomenon of word polycategory.

In some cases, NooJ might not transcribe the texts completely in only one run. Indeed grammatical sequences to be transformed can fail to be recognised if they include contractions or disjoint forms. It is therefore necessary to repeat the transcription of the text obtained at the first step. The transcription produces a new text without lemmas. A new lexical analysis needs to be performed.

The question of priority levels for morphological graphs is difficult. We must choose between "all at level minus one" that implies a plethora of answers and the repartition of morphological graphs between several levels so as not to have too many answers. But we might lose the good one in the process! New disambiguation grammars are needed, which will be ground for further research.

We would like to extend our warmest thanks to M. Silberztein who has, over the past twelve months, improved and developed the functionalities that we are using. Without his kind and patient collaboration and the quick help he gave us when requested, this work could not have been successful. What is extraordinary about working with the NooJ platform is the constant assistance provided by M. Silberztein, who is always devising new functionalities and making his software evolve to meet the needs of its users.

## References

Andersen et al, 2011. Methodological and Historical Dimensions of Corpus Linguistics. Vol. 6: Rayson, Hoffmann & Leech Ed. <www.helsinki.fi/varieng/journal/volumes/06>

Baron, A., Rayson, P. 2008. VARD2: A tool for dealing with spelling variation in historical corpora. In: Postgraduate Conference in Corpus Linguistics, 2008-05-22, Aston University, Birmingham.

Bauer, L. 1983. English Word Formation. Cambridge University Press.

Hogg, R. M., N. F. Blake, et al. 1992. The Cambridge History of the English Language. Vol. III. Cambridge: CUP.

Mesfar, S. 2006. Tutoriel: NooJ et la traduction. www.nooj4nlp.net/NooJ_Trad.pdf

Millward, C. M. 1996. A Biography of the English Language. Boston: Thomson Learning.

Pignot H., Piton O. 2010. Language Processing of 17th Century English and Creation of a Diachronic NooJ Dictionary, in Tamás Váradi, Judit Kuti and Max Silberztein eds., Applications of Finite-State Language Processing -- Selected Papers from the 2008 International NooJ Conference. Newcastle upon Tyne: Cambridge Scholars Publishing, 197-210.

Piton, O., Pignot, H. 2009. "Étude d'un corpus de textes de voyageurs anglais du 17e siècle, et aide à la transcription en anglais moderne". 6èmes JLC, Lorient. <http://www.licorn-ubs.com/jlc6.html>

Pignot, H., Piton, O. 2011. Mary Astell's Words in A Serious Proposal to the Ladies (part I), a Lexicographic Inquiry with NooJ, Proceedings of the 2010 NooJ Conference, Komotini, 232-44.

Scragg, D.G. 1974. A History of English Spelling. Manchester: MUP.Silberztein, M. 2005. "NooJ's dictionaries", Proceedings of the 2nd Language & Technology Conference, April 21-23, 2005, Poznán, Poland, Zygmunt Vetulani (ed.).

Silberztein, M. 2006. "NooJ's Linguistic Annotation Engine". In: Koeva, S., Maurel D., Silberztein M. (eds), INTEX/NooJ pour le Traitement Automatique des Langues. Cahiers de la MSH Ledoux. Presses Universitaires de Franche-Comté. 9-26.

Silberztein, M. 2007. "An Alternative Approach to Tagging". Invited paper. In: Proceedings of NLDB 2007. LNCS series. Berlin: Springer. 1-11.

Silberztein M. 2007. Complex Annotations with NooJ. In 2007 International NooJ Conference. Newcastle upon Tyne: Cambridge Scholars Publishing, 214-227.

Silberztein, M. NooJ manual. <http://www.nooj4nlp.net/pages/references.html>.

Visser, F. Th. 1963-73. An Historical Syntax of the English Language. Leiden: Brill.